\documentclass[runningheads]{llncs}
\usepackage{graphicx}
\usepackage{gensymb}
\usepackage{cite}
\usepackage{amsmath,amssymb,amsfonts}
\usepackage{algorithmicx}
\usepackage{textcomp}
\usepackage{xcolor}
\usepackage{multirow}
\usepackage{booktabs}
\usepackage{subcaption}
\usepackage{hyperref}
\usepackage{todonotes}
\usepackage{color,soul}
\hypersetup{
colorlinks = true,
linkcolor = red,
anchorcolor = blue,
citecolor = green,
filecolor = magenta,
urlcolor = cyan
}

\begin{document}
\title{ARTSeg: Employing Attention for Thermal images Semantic Segmentation}
\author{Farzeen Munir \and
Shoaib Azam  \and
Unse Fatima \and
Moongu Jeon }
\authorrunning{F. Munir et al.}
\institute{ School of Electrical Engineering and Computer Science \\
Gwangju Institute of Science and Technology, South Korea\\
\email{(farzeen.munir, shoaibazam, mgjeon)@gist.ac.kr, unse.fatima@gm.gist.ac.kr}}

\maketitle              % typeset the header of the contribution
\begin{abstract}

% Perception of the environment forms a critical backbone of an autonomous stack. Indeed, robust perception is integral to the safe driving of autonomous vehicles under different environmental conditions. For example, the visibility of RGB cameras is greatly affected during the night and low illumination conditions. A thermal camera proves to be an essential sensor in sensor suits of autonomous vehicles, using heat information to provide a more comprehensive, and therefore robust, perception.  This phenomenon of thermal cameras benefits semantic segmentation, which plays a pivotal role in the vision of autonomous vehicles. In this work, an attention Recurrent Convolution Neural Network (RCNN) based encoder-decoder architecture is proposed for thermal image segmentation. The main contribution is the design of encoder-decoder architecture, which employ units of RCNN for each encoder and decoder block. Furthermore, additive attention is employed in the decoder module to retain high-resolution features and improve the localization of features. The proposed method is evaluated on a public dataset, and efficacy is shown by calculating the mean Intersection over the union. 

The research advancements have made the neural network algorithms deployed in the autonomous vehicle to perceive the surrounding. The standard exteroceptive sensors that are utilized for the perception of the environment are cameras and Lidar. Therefore, the neural network algorithms developed using these exteroceptive sensors have provided the necessary solution for the autonomous vehicle's perception. One major drawback of these exteroceptive sensors is their operability in adverse weather conditions, for instance, low illumination and night conditions. The useability and affordability of thermal cameras in the sensor suite of the autonomous vehicle provide the necessary improvement in the autonomous vehicle's perception in adverse weather conditions. 
The semantics of the environment benefits the robust perception, which can be achieved by segmenting different objects in the scene. In this work, we have employed the thermal camera for semantic segmentation. We have designed an attention-based Recurrent Convolution Network (RCNN) encoder-decoder architecture named ARTSeg for thermal semantic segmentation. The main contribution of this work is the design of encoder-decoder architecture, which employ units of RCNN for each encoder and decoder block. Furthermore, additive attention is employed in the decoder module to retain high-resolution features and improve the localization of features. The efficacy of the proposed method is evaluated on the available public dataset, showing better performance with other state-of-the-art methods in mean intersection over union (IoU).

\end{abstract}
\section{Introduction}
The last three decades have shown significant technological advancements that reflects the development of efficient sensors. The applicability of these sensors in the autonomous vehicle ensures its safe and secure services to the urban environment, as indicated by SOTIF-ISO/PAS-21448\footnote{https://www.daimler.com/innovation/case/autonomous/safety-first-for-automated-driving-2.htm}. Keeping safety as a priority in autonomous vehicles, perception of the environment play a critical role along with localization, planning and control module \cite{Azam2020} \cite{Munir2018} \cite{shoaib2020}. In the context of perception, the most common exteroceptive sensors that are being deployed are cameras (visible spectrum) and Lidar. The utilization of these sensor modalities provides the necessary perception for the autonomous vehicle. However, these sensors have limitations in adverse weather conditions at night and low illumination environments. For instance, cameras (visible spectrum) are operated in the visible spectrum domain and provide an in-depth understanding of the environment. Still, environmental conditions like sun-glare, low illumination affect the camera result, thus yielding low performance in perception algorithms.
On the other hand, as a surrogate to $2$D information from cameras, Lidar gives the $3$D information about the environment. Lidar provides 3D information of the environment by making a point cloud map by projecting nearly thousands of laser beams to the environment. Besides, Lidar effectiveness in providing the detailed $3$D representation of the environment, its expensive cost and limitation in adverse weather conditions are the prime concern of its detriment. On the other hand, thermal cameras, contrary to cameras (visible-spectrum) and Lidar, enable the perception algorithms to be utilized in adverse weather conditions, such as at night or low illumination environmental conditions \cite{Rosique2019}.
\par
Thermal cameras operate in the infrared domain and capture the infrared radiation emitted by the different entities in the environment having the temperature above absolute zero \cite{Vollmer2017}. This property of the thermal camera makes it an optimal solution to be included in the sensor suite of the autonomous vehicle for the perception of the environment in low illumination and night conditions. Furthermore, besides thermal cameras applicability in adverse weather conditions, the affordability of thermal cameras gives the potential to be utilized in different perception tasks for the autonomous vehicle. These perception tasks involve different computer vision applications such as object detection \cite{Hwang2015} \cite {Xu2017}, visual tracking \cite{Li2018} and person re-identification \cite{Wu2017}. In order to determine the semantics of environment for the scene understanding for the autonomous vehicle, semantic segmentation plays a vital role in the perception of the autonomous vehicle. The capacity of thermal cameras to operate at night and low illumination conditions motivate to use of the thermal camera for semantic segmentation and investigate the semantic segmentation problem using the thermal cameras. 
\par 
In computer vision, semantic segmentation provides an in-depth understanding of the scene by employing a semantic label to each pixel in the image. Despite traditional machine learning approaches used to tackle semantic segmentation problems, deep learning approaches gain unprecedented success. For instance, most of these deep learning techniques (convolutional neural networks) are applied using the RGB images for semantic segmentation.  Thermal cameras provide the grey-scale image that benefits the neural networks employed on thermal images for different computer vision tasks. 
\par
This work explores the thermal image segmentation problem by designing a novel encoder-decoder architecture called ARTSeg. The encoder consists of a residual recurrent convolution block accompanied by a max-pooling layer. The encoder outputs a latent representation of features which is fed to the decoder. The decoder incorporates additive attention from the residual connections from the encoder. The ARTSeg is evaluated on a public dataset and compared with state-of-the-art methods. The main contributions of this work are as follows:
\begin{enumerate}
    \item We have designed a novel encoder-decoder architecture using a residual recurrent convolution neural network.
    \item We have employed the additive attention mechanism to enhance the localization of encoded features. Further, no post-processing steps are used because of utilizing the attention for the thermal image segmentation.
    \item The proposed method's efficacy is performed on available public datasets shows better performance in terms of average accuracy and intersection over the union in contrast to state-of-the-art methods.
\end{enumerate}
The rest of the paper is organized as follows: Section 2 explains the related work. The proposed method is described in the Section 3. Section 4 illustrate the experimentation and results for the proposed method. Finally, section 5 concludes the paper.

\section{Related Work}
The concept of semantic segmentation plays a critical role in the robust perception of the environment for autonomous driving. Deep neural networks, especially convolution neural networks, are mostly used for semantic segmentation using RGB cameras as a sensor modality. The early state-of-the-art method that is employed for the semantic segmentation is Fully Convolutional Network (FCN) proposed by \cite{a}. In FCN, the fully connected layers are replaced with the convolutional layers to output the full resolution maps for the semantic segmentation using the backbone network of VGG16 \cite{vgg}, and GoogleNet \cite{c} architectures, respectively.
\par 
In literature, the encoder-decoder architecture is also used for semantic segmentation. In the encoder-decoder architecture, the features are encoded to give the latent representation and then decoded to provide the object details and spatial resolution. \cite{segnet} have proposed the encoder-decoder architecture named SegNet. The encoder network consists of convolutional layers adopted from the VGG16 \cite{vgg} network to learn the encoded representation followed by batch-normalization, rectified linear unit (ReLU) units and max-pooling layers. The decoder follows the same encoder architecture for upsampling the encoded features for semantic segmentation. Similarly, using the encoder-decoder architecture, the encoder of the network is improved by using image pyramids \cite{Chen2017}\cite{Chen2016}, conditional random fields \cite{Chen2017}, spatial pyramid pooling \cite{Chen2017} and atrous convolution \cite{Papandreou2017} \cite{Wang2018}. There is a trade-off between accuracy and speed; in literature, some research is focused on improving the inference speed using the encoder-decoder architecture; for instance, \cite{e} have proposed the encoder-decoder architecture ENet that is optimized for the fast inference speed for semantic segmentation. In order to retain the feature generalization, skip connections have been introduced for the semantic segmentation as proposed by UNet \cite{d}.
\par 
Besides, using the single sensor modality, for instance, RGB images, in literature, the fusion of thermal and visible-spectrum RGB domain is also investigated for semantic segmentation. \cite{j} have proposed FuseNet for the semantic segmentation by incorporating the fusion of visible-spectrum RGB domain with the thermal domain. Similarly, MFNet \cite{Qishen} and RTFNet \cite{rtfnet} networks follows the encoder-decoder architecture for the semantic segmentation using the fusion of RGB and thermal data. \cite{fuseseg} have proposed the FuseSeg using the Bayesian fusion theory for the semantic segmentation using both RGB and thermal data. In addition to the fusion techniques, some research is focused on multi-spectral domain adaptation \cite{Kim2021} \cite{Tsai2018} \cite{Kim2020}. 
\par
In contrast to the literature review, we have designed a novel encoder-decoder architecture using the residual recurrent convolution layers followed by an attention mechanism to retain the feature generalization for semantic segmentation. Furthermore, we have explicitly used the thermal data for the semantic segmentation without employing the fusion or domain adaption techniques in this work. 
% \usepackage{booktabs}
% \usepackage{multirow}
% \usepackage{graphicx}
\begin{figure*}[t]
\centering
\includegraphics[width=\textwidth]{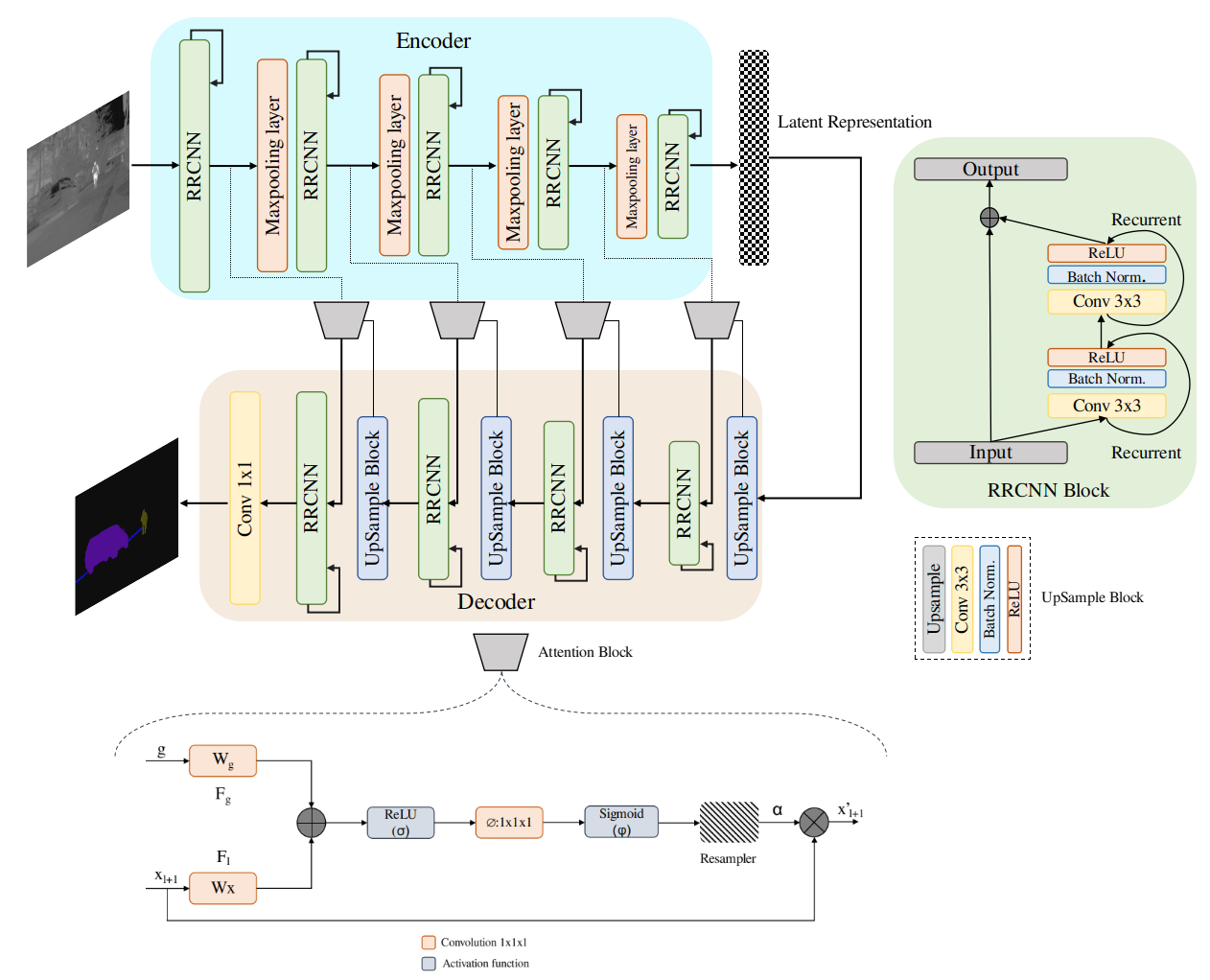}
\caption{ The overall framework of ARTSeg Network consists of an encoder-decoder architecture. }
\label{framework}
\end{figure*}
\begin{table}[!b]
\centering
\caption{The detailed architecture of the proposed ARTSeg}
\label{layer}
\resizebox{5cm}{!}{%
\begin{tabular}{@{}l|l|c@{}}
\toprule
                         & Layer            & Output size \\ \midrule
                         & Input data       &             \\ \midrule
\multirow{8}{*}{Encoder} & RRCNN-1 block    & 32*256*256  \\
                         & Max pooling      & 32*128*128  \\
                         & RRCNN-2 block    & 64*128*128  \\
                         & Max pooling      & 64*64*64    \\
                         & RRCNN-3 block    & 128*64*64   \\
                         & Max pooling      & 128*32*32   \\
                         & RRCNN-4 block    & 256*32*32   \\
                         & Max pooling      & 256*8*8     \\
                         & RCNN-5 block     & 256*8*8     \\ \midrule
\multirow{9}{*}{Decoder} & Up-Block         & 128*32*32   \\
                         & Attention Module & 256*32*32   \\
                         & RRCNN-4 block    & 128*32*32   \\
                         & Up-Block         & 128*64*64   \\
                         & Attention Module & 256*64*64   \\
                         & RRCNN-3 block    & 128*64*64   \\
                         & Up-Block         & 64*128*128  \\
                         & Attention Module & 128*128*128 \\
                         & RRCNN-2 block    & 64*128*128  \\
                         & Up-Block         & 32*256*256  \\
                         & Attention Module & 64*256*256  \\
                         & RRCNN-1 Block    & 32*256*256  \\ \midrule
Output                   & Conv 1X1         & 5*256*256   \\ \bottomrule
\end{tabular}%
}
\end{table}

\section{Methods}
This section presents the proposed method for the thermal image segmentation, as illustrated in Fig.\ref{framework}. The proposed method is composed of encoder-decoder architecture. The encoder module of the proposed method contains the residual recurrent convolution block (RRCNN) followed by the max-pooling layer.  The residual recurrent convolutional block is comprised of recurrent convolutional layers and skip connections from the encoder to decoder module. The encoder module of the architecture receives the thermal image as input and generates the encoded feature representation for the input image. Suppose $x_l$ represent the input image at the $l^{th}$ layer of the residual recurrent convolutional block, and a pixel is located at $(u,v)$ at the input image on the $p$ feature map of the recurrent convolution layer, then the output $o_{uvp}$ is expressed in Eq.\ref{1}.

\begin{equation}
% \begin{aligned}
 o_{uvp}^{l}= (w_p^f)^T * x_l^f(u,v)(t) + (w_p^r)^T * x_l^r(u,v)(t-1) + b_p,
\label{1}
% \end{aligned}
\end{equation}

where $w_p^f$ and $w_p^r$ are the weights of convolutional and recurrent convolutional layer respectively and $b_p$ is the bias of the network. The inputs to the convolution and $l^{th}$ layer recurrent convolutional layer are represented as $x_l^f(u,v)(t)$ and $x_l^r(u,v)(t-1)$ respectively. In Fig.\ref{1} the recurrent convolution layer are represented in the RRCNN block with the ``arrow" indicating the recurrence. The output of the recurrent convolutional layer is batch-normalized, and fed to the rectified linear activation function (ReLU) as shown in Eq.\ref{2}.

\begin{equation}
% \begin{aligned}
 \mathfrak{F}(\mathbf{x_l},\boldsymbol{w_l})=f(o_{uvp}^{l}(t)))=max(0,o_{uvp}^{l}(t)),
\label{2}
% \end{aligned}
\end{equation}
$\mathfrak{F}(\mathbf{x_l},\boldsymbol{w_l})$ represents the outcome of the activation function from the $l^{th}$ layer recurrent convolutional network. This is summed with the input $x_l$ in residual to give us the output of RRCNN block as expressed as Eq.\ref{3}.
\begin{equation}
% \begin{aligned}
 x_{l+1}=x_l+\mathfrak{F}(\mathbf{x_l},\boldsymbol{w_l}),
\label{3}
% \end{aligned}
\end{equation}
where $x_{l+1}$ represent the outcome from the RRCNN block.
\par 
The decoder module follows the same architecture of the encoder module with the addition of Upsample block and additive attention in the skip connections. The decoder includes the same RRCNN block as the encoder module, but the RRCNN module is used to upsampling the encoded features. Fig.\ref{1} shows the decoder module with attention block. The output of the attention is calculated by the element-wise multiplication of attention coefficient and input feature maps from as shown in Eq.\ref{4}.
\begin{equation}
% \begin{aligned}
 x_{l+1}' = x_{l+1} \cdot \alpha,
\label{4}
% \end{aligned}
\end{equation}
 The purpose of the attention module is to focus on the salient region for the thermal image segmentation. Mathematically, the attention of the network is given by Eq.\ref{5}. The term g represents the vector taken from the lowest layer of the network. The detailed architecture is presented in Table-\ref{layer}.

\begin{equation}
% \hspace{-2.5cm}
\begin{aligned}
 & s^l_{att}=\phi^T (\sigma (W^T_xx^i_{l+1}+W^T_gg^i+b_g))+b_\phi, \\
 & \alpha ^l=\varphi(s^l_{att}(x^i_{l+1},g^i;\Theta_{att} )), 
 \label{5}
\end{aligned}
\end{equation}

where $\sigma$ and $\varphi$ represent ReLU and sigmoid activation function respectively. The $W_x$ and $W_g$ shows the linear transformation for the attention network. In addition, the Upsample block includes an upsampling layer followed by convolutional layer, batch normalization and the ReLU activation unit. The semantic segmentation output is obtained by placing the final convolution layer of kernel size $1 \times 1$ at the final RRCNN block in the decoder layer. 
% \usepackage{booktabs}
% \usepackage{graphicx}
% \usepackage{booktabs}
% \usepackage{graphicx}
\begin{table}[!b]
\centering
\caption{The number of images in training and testing split of the dataset.}
\label{dataset}
\resizebox{7cm}{!}{%
\begin{tabular}{@{}l|l|c@{}}
\toprule
Dataset         & Training set             & Testing set \\ \midrule
Day \& Night time & \multicolumn{1}{c|}{784} & 393         \\
Day Time        &         \hspace{0.7cm}  -               & 205         \\
Night Time      &           \hspace{0.7cm}  -               & 188         \\ \bottomrule
\end{tabular}%
}
\end{table}
% \begin{table}[t]
% \centering
% \caption{The number of images in training and testing split of the dataset.}
% \label{dataset}
% \resizebox{9cm}{!}{%
% \begin{tabular}{@{}l|c|c@{}}
% \toprule
% Dataset      & Training set & Testing set \\ \midrule
% day\&night time & 784    & 393  \\
% day time         &       & 205  \\
% night time       &       & 188  \\ \bottomrule
% \end{tabular}%
% }
% \end{table}
\section{Experimentation and Results}
This section explains the experimentation and results of the proposed method. The performance of the proposed method is evaluated on the available public dataset\cite{Qishen}. The efficacy of the proposed method is compared with the state-of-the-art methods for thermal image semantic segmentation. For the evaluation, we have adopted the standard metrics: Average class accuracy and mean Intersection over Union (IoU). The following section discusses the details of experimentation and results.
\subsection{Thermal Semantic Segmentation Dataset}
We have utilized the available public dataset for thermal image segmentation provided by \cite{Qishen}. The dataset is comprised of thermal and visible-spectrum RGB images. The dataset is collected using the InfRec R500 thermal camera, simultaneously capturing the thermal spectrum and visible spectrum images. The range of the thermal spectrum for collecting the images is $814 \mu m$, giving the images of resolution $480\times 640$. However, the field of view of the visible-spectrum image and the thermal image is not identical. The visible spectrum images have a horizontal field of view of $100^{\circ}$; besides, the thermal spectrum images have a horizontal field of view of $32^{\circ}$. In order to align the thermal and RGB spectrum images, the RGB images are cropped and resized. The dataset provides the semantic labels for the urban environment and is classified into nine classes: bicycle, person, car, curve, car\_stop, color\_cone, guardrail, and background. The dataset has 1569 images pairs of RGB and thermal altogether. There are 820 image pairs recorded during the daytime and 749 recorded at nighttime. The dataset is split into the training set and testing set. The testing set is further split into the day and nighttime groups. The number of images in each group is shown in Table \ref{dataset}.  
%

% \usepackage{booktabs}
% \usepackage{graphicx}

\subsection{Training details}
The proposed method ArtSeg is implemented using Pytorch deep learning library. In training the proposed method, no pre-processing step is performed on the input images. The proposed network ArtSeg is trained from scratch in an end-to-end manner and does not employ any pre-trained weights. The network is trained for a total of 500 epochs on Nvidia RTX $3090$ having $24GB$ memory. The cross-entropy loss function is used for training the network given by Eq.\ref{loss}. 
\begin{equation}
% \hspace{-2.5cm}
\begin{aligned}
Loss=-\frac{1}{M}\sum_{j=1}^{M}\sum_{c=1}^{C}S_{c,j} \ln (\hat{S}_{c,j})
 \label{loss}
\end{aligned}
\end{equation}
where $M$ is the total number of pixels in the ground truth label image, $C$ is the number of classes, in this case, $9$. $S_{c,j}$ denotes the ground truth class label of each pixel, and $\hat{S}_{c,j}$ denotes the predicted class of each pixel.  We have used Adam optimizer for training the ARTSeg network with the weight decay of $1\times 10^{-4}$; epsilon of $1\times 10^{-8}$ and learning rate of $5\times10^{-4}$. The learning rate schedule policy given by the Eq.\ref{eq1} is used to update the learning rate. 
\begin{equation}
% \hspace{-2.1cm}
% \begin{aligned}
LR=LR_{initial} \times (\frac{1-epoch}{epochs_{total}})^{p},
\label{eq1}
% \end{aligned}
\end{equation}
The $p$ is set to $0.9$. The network is trained for 100 epochs. All the parameter values are chosen empirically through grid search.  These parameters are kept constant in all the experiments. Moreover, training data is augmented using the flip technique. 
\begin{table}[t]
\centering
\caption{ Comparison of the evaluation results of ARTSeg with other state-of-the-art methods on the test set consisting of Day and Night time images. The Class Accuracy (\%), Avg.Acc (\%) and $IOU$ (\%) are the evaluation metrics. "-" indicates that the value is not available. }
\label{com}
\resizebox{12cm}{!}{%
\begin{tabular}{@{}l|ccccccccc|cc@{}}
\toprule
Model       & Background & Car   & Pedestrian & Bike & Curve & Car stop & Guardrail & Color cone & Bump  & Avg.Acc & $IOU$ \\ \midrule
SegNet \cite{Qishen}   & 96.90 & 83.30 & 72.10 & 76.80 & 58.30 & 31.90 & 0.00  & 0.00  & 63.90 & 53.70 & 58.30 \\
SegNet\_4ch \cite{Qishen}  & 96.10      & 89.00 & 82.30      & 0.00 & 61.40 & 21.70    & 0.00      & 0.00       & \textbf{86.70} & 48.60   & 50.40 \\
ENet \cite{Qishen}     & 88.50 & 58.60 & 42.70 & 24.70 & 30.10 & 18.10 & 0.30  & 45.80 & 23.00 & 37.00 & 44.90 \\
MFNet \cite{Qishen}   & 96.80 & 82.90 & 85.20 & 74.20 & 61.50 & 27.30 & 0.00  & 60.70 & 43.30 & 59.10 & 64.90 \\
FuseSeg \cite{fuseseg} & -     & 93.10 & 81.40 & 78.50 & 68.40 & 29.10 & 63.70 & 55.80 & 66.64 & 70.60 & 54.50 \\
RTFNet \cite{rtfnet} & -     & 93.00 & 79.30 & 76.80 & 60.70 & 38.50 & 0.00  & 45.50 & 74.70 & 63.10 & 53.20 \\ \midrule
\textbf{ARTSeg (Our)}     & \textbf{97.10} & \textbf{94.76} & \textbf{86.66} & \textbf{79.20} & \textbf{71.25} & \textbf{49.69} & \textbf{65.21} &\textbf{ 58.11}  & 64.28 & \textbf{74.03} & \textbf{68.80} \\ \bottomrule
\end{tabular}%
}
\end{table}
\subsection{Evaluation metrics}
In order to evaluate the performance of the proposed network, two evaluation metrics are selected. The Average class accuracy and mean Intersection over Union (IoU). The Average class accuracy is expressed as 
\begin{equation}
% \hspace{-5.1cm}
% \begin{aligned}
 Avg.Acc=\frac{1}{N}\sum_{i=1}^{N}\frac{TP_i}{TP_i+FN_i},\\
\label{aa}
% \end{aligned}
\end{equation}
where $N$ is number of classes.$TP$ is true positive rate ($TP_i=\sum_{m=1}^{M}p_{ii}^m$) and $FN$ is false negative rate(${FN}_i=\sum_{m=1}^{M}\sum_{j=1,j \neq i}^{N}p_{ij}^m$).
The $IoU$ is expressed as 
\begin{equation}
% \hspace{-5.1cm}
% \begin{aligned}
 IoU=\frac{1}{N}\sum_{i=1}^{N}\frac{TP_i}{TP_i+FN_i+ FP_i},\\
\label{miou}
% \end{aligned}
\end{equation}
here $FP$ is define as false positive rate(${FP}_i=\sum_{m=1}^{M}\sum_{j=1,j \neq i}^{N}p_{ji}^m$). The $p_{ii}$ corresponds to the correct number of pixels classified for class $i$ having the same class $i$ in the frame $m$, whereas $p_{ji}$ represents the number incorrectly classified pixel of class $j$ as class $i$ in the frame $m$. Similarly, the number of incorrectly classified pixel of class $i$ as class $j$ is represented by $p_{ij}$ in the frame $m$.
\subsection{Results}
\begin{figure*}[t]
\centering
\includegraphics[width=\textwidth]{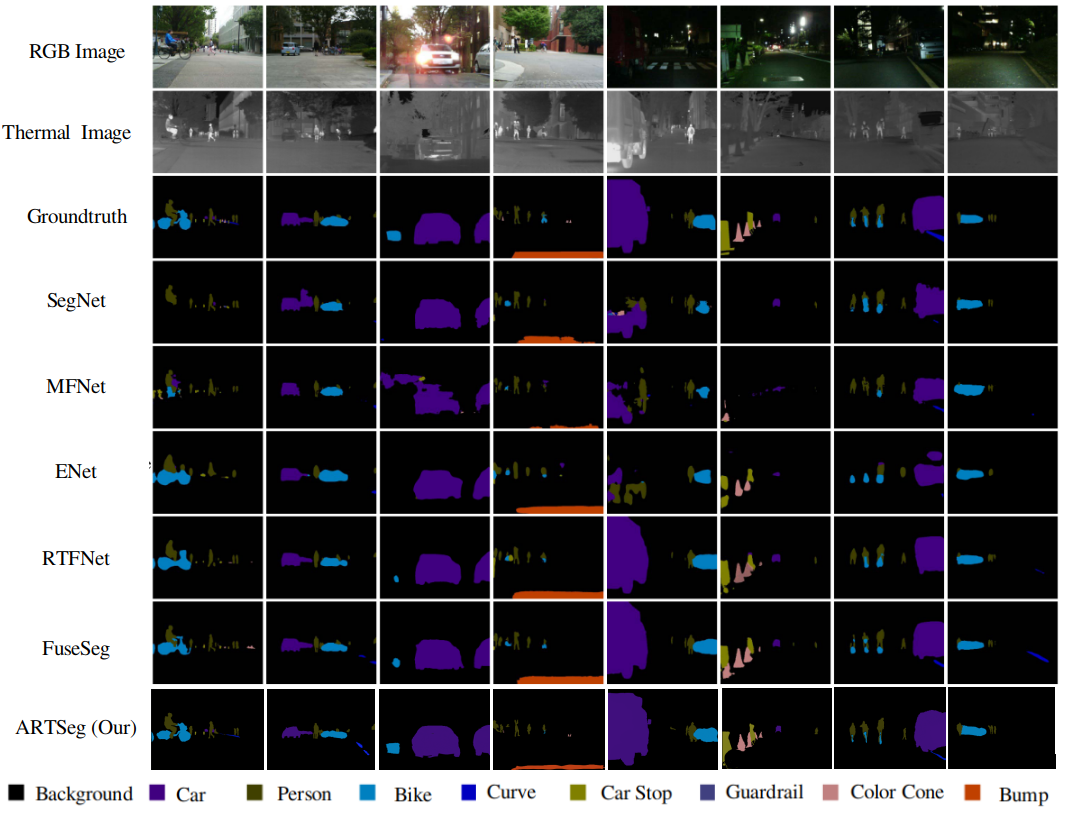}
\caption{ The qualitative comparison between ARTSeg and other state-of-the-art methods for semantic segmentation on thermal image dataset.   }
\label{results}
\end{figure*}
\begin{table}[!b]
\centering
\caption{Quantitative analysis of ARTSeg with other state-of-the-art methods for day time and night time test set. The evaluation is performed using Avg.acc (\%) and $IoU$ (\%) metrics. }
\label{Sep}
\resizebox{8cm}{!}{%
\begin{tabular}{@{}l|cc|cc@{}}
\toprule
Methods     & \multicolumn{2}{c|}{Day Time} & \multicolumn{2}{c|}{Night Time} \\ \midrule
            & Avg.Acc(\%)        & $IOU$(\%)       & Avg.Acc(\%)          & $IOU$(\%)         \\ \midrule
SegNet \cite{Qishen}     & 46.10          & 48.80       & 54.00           & 55.20        \\
SegNet\_4ch \cite{Qishen} & 50.50          & 51.10       & 50.50           & 51.10        \\
MFNet  \cite{Qishen}     & 47.70          & 57.40       & 63.50           & 62.10        \\
FuseSeg \cite{fuseseg}     & 62.10          & 47.80       & 67.30           & 54.60        \\
RTFNet\cite{rtfnet}      & 60.00          & 45.80       & 60.70           & 54.80        \\
\textbf{ARTSeg (Our)}         & \textbf{65.58}          & \textbf{64.08}       &\textbf{ 70.85 }          & \textbf{65.56}        \\ \bottomrule
\end{tabular}%
}
\end{table}
The thermal image semantic segmentation dataset is benchmarked on existing techniques, including MFRNet \cite{Qishen}, FuseSeg\cite{fuseseg}, SegNet\cite{segnet} and RTFNet\cite{rtfnet}. Image segmentation using thermal images is an emerging research direction in contrast to RGB image segmentation; given this, a few techniques have been employed in the literature for thermal image segmentation. The MFRNet, FuseSeg and RTFNet have fused the information from the visible spectrum and thermal spectrum images.  However, SegNet uses only thermal images for predicting semantic segmentation. The proposed network ARTSeg is trained on only thermal images. The evaluation of ARTSeg in comparison to the algorithms as mentioned earlier is shown in Table \ref{com}. The Table shows the Accuracy measure of each class and Avg.Acc and $IoU$ over all classes for both day and night time. However, Table \ref{Sep} shows the evaluation of the methods for daytime and nighttime separately for all classes.  ARTSeg has outperformed all existing methods and achieved an improvement $3.43\%$ on Avg.Acc and $3.9\%$ in $IoU$ scores in comparison to FuseSeg. Fig.\ref{results} manifests the qualitative results of the proposed ARTSeg algorithm with other methods. In addition,  we have also investigated the effect of using different backbone networks to extract the features for the encoder of the ARTSeg network. 

\begin{table}[t]
\centering
\caption{Quantitive analysis of ARTSeg with different backbone networks. The evaluation is performed using Avg.acc (\%) and $IOU$ (\%) metrics on day and night time test set.}
\label{back}
\resizebox{8cm}{!}{%
\begin{tabular}{@{}l|cc|cc@{}}
\toprule
Methods            & \multicolumn{2}{c|}{DayTime} & \multicolumn{2}{c|}{NightTime} \\ \midrule
                   & Avg.Acc(\%)         & $IOU$ (\%)       & Avg.Acc(\%)          & $IOU$(\%)         \\ \midrule
ARTSeg-VGG16       &   54.21       & 46.80       & 59.12           & 48.20        \\
ARTSeg-ResNet-18   &   57.50       & 47.60     & 61.89          & 51.10        \\
ARTSeg-ResNet-50   & 63.25       & 55.40       & 65.50           & 57.23        \\
ARTSeg-MobileNetv2 & 59.45       & 49.89       & 60.10          & 50.80        \\
ARTSeg-DenseNet    & 58.41         & 49.50       & 59.45           & 49.34        \\
ARTSeg-ShuffleNet  & 62.58          & 58.18       & 66.85           & 59.56        \\ 
\textbf{ARTSeg (Our) }        & \textbf{65.58}          & \textbf{64.08}       & \textbf{70.85 }          & \textbf{ 65.56 } \\  \bottomrule
\end{tabular}%
}
\end{table}

We utilize six different backbone network including VGG16 \cite{vgg}, ResNet-18 \cite{resnet}, ResNet-50 \cite{resnet}, MobileNetV2 \cite{mobilenet}, ShuffleNet \cite{shufflenet} and DenseNet \cite{dense}. The thermal image is input to the backbone network. The encoded features are then passed to the decoder module. The backbone networks are trained with pre-trained weights. The Table \ref{back} shows the quantitative results with the backbone networks. The results using different backbone networks in the encoder module do not show any significant improvement compared to ARTSeg.

% \usepackage{booktabs}
% \usepackage{graphicx}

% \usepackage{booktabs}
% \usepackage{graphicx}

% \usepackage{booktabs}
% \usepackage{graphicx}

\section{Conclusion}
In this research article, we proposed ARTSeg, a novel encoder-decoder architecture for thermal image semantic segmentation. The ARTSeg introduces residual recurrent convolution block in the encoder, followed by a decoder that utilizes additive attention in skip connection to refine full resolution detection. The ARTSeg is evaluated on the public dataset in terms of Avg.Acc and $IoU$. In comparison to other state-of-the-art methods, ARTSeg has a higher Avg.Acc ($74.03\%$) and $IoU$ ($68.80\%$).
\par
The inclusion of the thermal camera in the autonomous driving stack provide help to understand the surroundings in low illumination conditions. In future work, we aim to fuse information from the thermal camera and Lidar to improve the autonomous vehicle's perception.

\section*{Acknowledgement}
This work was partly supported by the ICT R$\&$D program of MSIP/IITP (2014-3-00077, Development of global multitarget tracking and event prediction techniques based on real-time large-scale video analysis), National Research Foundation of Korea (NRF) grant funded by the Korea Government (MSIT) (No. 2019R1A2C2087489), Ministry of Culture, Sports and Tourism (MCST), and Korea Creative Content Agency (KOCCA) in the Culture Technology (CT) Research \& Development (R2020070004) Program 2021.


\begin{thebibliography}{99}
\bibitem{Azam2020} Azam, S., Munir, F., Sheri, A.M., Kim, J. and Jeon, M., 2020. System, design and experimental validation of autonomous vehicle in an unconstrained environment. Sensors, 20(21), p.5999.
\bibitem{Munir2018}Munir, F., Azam, S., Hussain, M.I., Sheri, A.M. and Jeon, M., 2018, October. Autonomous vehicle: The architecture aspect of self driving car. In Proceedings of the 2018 International Conference on Sensors, Signal and Image Processing (pp. 1-5).
\bibitem{shoaib2020}Azam, S., Munir, F. and Jeon, M., 2020. Dynamic Control System Design for Autonomous Car. In VEHITS (pp. 456-463).
\bibitem{Rosique2019}Rosique, F., Navarro, P.J., Fernández, C. and Padilla, A., 2019. A systematic review of perception system and simulators for autonomous vehicles research. Sensors, 19(3), p.648.
\bibitem{a}Long, Jonathan, Evan Shelhamer, and Trevor Darrell. "Fully convolutional networks for semantic segmentation." In Proceedings of the IEEE conference on computer vision and pattern recognition, pp. 3431-3440. 2015.
\bibitem{Vollmer2017}Vollmer, M. and Möllmann, K.P., 2017. Infrared thermal imaging: fundamentals, research and applications. John Wiley \& Sons.
\bibitem{Hwang2015}Hwang, S., Park, J., Kim, N., Choi, Y. and So Kweon, I., 2015. Multispectral pedestrian detection: Benchmark dataset and baseline. In Proceedings of the IEEE conference on computer vision and pattern recognition (pp. 1037-1045).
\bibitem{Xu2017}Xu, D., Ouyang, W., Ricci, E., Wang, X. and Sebe, N., 2017. Learning cross-modal deep representations for robust pedestrian detection. In Proceedings of the IEEE conference on computer vision and pattern recognition (pp. 5363-5371).
\bibitem{Li2018}Li, C., Zhu, C., Huang, Y., Tang, J. and Wang, L., 2018. Cross-modal ranking with soft consistency and noisy labels for robust rgb-t tracking. In Proceedings of the European Conference on Computer Vision (ECCV) (pp. 808-823).
\bibitem{Wu2017}Wu, A., Zheng, W.S., Yu, H.X., Gong, S. and Lai, J., 2017. RGB-infrared cross-modality person re-identification. In Proceedings of the IEEE international conference on computer vision (pp. 5380-5389).
% \bibitem{b}Simonyan, Karen, and Andrew Zisserman. "Very deep convolutional networks for large-scale image recognition." arXiv preprint arXiv:1409.1556 (2014).
\bibitem{c}Szegedy, Christian, Wei Liu, Yangqing Jia, Pierre Sermanet, Scott Reed, Dragomir Anguelov, Dumitru Erhan, Vincent Vanhoucke, and Andrew Rabinovich. "Going deeper with convolutions." In Proceedings of the IEEE conference on computer vision and pattern recognition, pp. 1-9. 2015.
\bibitem{d}Ronneberger, Olaf, Philipp Fischer, and Thomas Brox. "U-net: Convolutional networks for biomedical image segmentation." In International Conference on Medical image computing and computer-assisted intervention, pp. 234-241. Springer, Cham, 2015.
\bibitem{e}Paszke, Adam, Abhishek Chaurasia, Sangpil Kim, and Eugenio Culurciello. "Enet: A deep neural network architecture for real-time semantic segmentation." arXiv preprint arXiv:1606.02147 (2016).
\bibitem{f}Song, Shuran, Samuel P. Lichtenberg, and Jianxiong Xiao. "Sun rgb-d: A rgb-d scene understanding benchmark suite." In Proceedings of the IEEE conference on computer vision and pattern recognition, pp. 567-576. 2015.
\bibitem{j}Hazirbas, Caner, Lingni Ma, Csaba Domokos, and Daniel Cremers. "Fusenet: Incorporating depth into semantic segmentation via fusion-based cnn architecture." In Asian conference on computer vision, pp. 213-228. Springer, Cham, 2016.
\bibitem{l}Yu, Changqian, Jingbo Wang, Chao Peng, Changxin Gao, Gang Yu, and Nong Sang. "Learning a discriminative feature network for semantic segmentation." In Proceedings of the IEEE conference on computer vision and pattern recognition, pp. 1857-1866. 2018.
\bibitem{m}Qiao, Yulong, Ziwei Wei, and Yan Zhao. "Thermal infrared pedestrian image segmentation using level set method." Sensors 17, no. 8 (2017): 1811.
\bibitem{n}Li, Chenglong, Wei Xia, Yan Yan, Bin Luo, and Jin Tang. "Segmenting objects in day and night: Edge-conditioned cnn for thermal image semantic segmentation." IEEE Transactions on Neural Networks and Learning Systems (2020).

\bibitem{Qishen}Ha, Qishen, Kohei Watanabe, Takumi Karasawa, Yoshitaka Ushiku, and Tatsuya Harada. "MFNet: Towards real-time semantic segmentation for autonomous vehicles with multi-spectral scenes." In 2017 IEEE/RSJ International Conference on Intelligent Robots and Systems (IROS), pp. 5108-5115. IEEE, 2017.
\bibitem{fuseseg}Sun, Yuxiang, Weixun Zuo, Peng Yun, Hengli Wang, and Ming Liu. "FuseSeg: semantic segmentation of urban scenes based on RGB and thermal data fusion." IEEE Transactions on Automation Science and Engineering (2020).
\bibitem{rtfnet}Sun, Yuxiang, Weixun Zuo, and Ming Liu. "Rtfnet: Rgb-thermal fusion network for semantic segmentation of urban scenes." IEEE Robotics and Automation Letters 4, no. 3 (2019): 2576-2583.
\bibitem{segnet}Badrinarayanan, Vijay, Alex Kendall, and Roberto Cipolla. "Segnet: A deep convolutional encoder-decoder architecture for image segmentation." IEEE transactions on pattern analysis and machine intelligence 39, no. 12 (2017): 2481-2495.
\bibitem{vgg}Simonyan, Karen, and Andrew Zisserman. ``Very deep convolutional networks for large-scale image recognition.`` arXiv preprint arXiv:1409.1556 (2014).
\bibitem{resnet} He, Kaiming, Xiangyu Zhang, Shaoqing Ren, and Jian Sun. ``Deep residual learning for image recognition.`` In Proceedings of the IEEE conference on computer vision and pattern recognition, pp. 770-778. 2016.
\bibitem{mobilenet} Sandler, Mark, Andrew Howard, Menglong Zhu, Andrey Zhmoginov, and Liang-Chieh Chen. ``Mobilenetv2: Inverted residuals and linear bottlenecks.`` In Proceedings of the IEEE conference on computer vision and pattern recognition, pp. 4510-4520. 2018.
\bibitem{shufflenet} Ma, Ningning, Xiangyu Zhang, Hai-Tao Zheng, and Jian Sun. ``Shufflenet v2: Practical guidelines for efficient cnn architecture design.`` In Proceedings of the European conference on computer vision (ECCV), pp. 116-131. 2018.
\bibitem{dense} Huang, Gao, Zhuang Liu, and Kilian Q. Weinberger. ``Densely connected convolutional networks. CoRR abs/1608.06993 (2016).`` arXiv preprint arXiv:1608.06993 (2016).

\bibitem{Chen2017}Chen, L.C., Papandreou, G., Kokkinos, I., Murphy, K. and Yuille, A.L., 2017. Deeplab: Semantic image segmentation with deep convolutional nets, atrous convolution, and fully connected crfs. IEEE transactions on pattern analysis and machine intelligence, 40(4), pp.834-848.

\bibitem{Chen2016}Chen, L.C., Yang, Y., Wang, J., Xu, W. and Yuille, A.L., 2016. Attention to scale: Scale-aware semantic image segmentation. In Proceedings of the IEEE conference on computer vision and pattern recognition (pp. 3640-3649).

\bibitem{Papandreou2017}Chen, L.C., Papandreou, G., Schroff, F. and Adam, H., 2017. Rethinking atrous convolution for semantic image segmentation. arXiv preprint arXiv:1706.05587.

\bibitem{Wang2018}Wang, P., Chen, P., Yuan, Y., Liu, D., Huang, Z., Hou, X. and Cottrell, G., 2018, March. Understanding convolution for semantic segmentation. In 2018 IEEE winter conference on applications of computer vision (WACV) (pp. 1451-1460). IEEE.

\bibitem{Kim2021}Kim, Y.H., Shin, U., Park, J. and Kweon, I.S., 2021. MS-UDA: Multi-Spectral Unsupervised Domain Adaptation for Thermal Image Semantic Segmentation. IEEE Robotics and Automation Letters.

\bibitem{Tsai2018}Tsai, Y.H., Hung, W.C., Schulter, S., Sohn, K., Yang, M.H. and Chandraker, M., 2018. Learning to adapt structured output space for semantic segmentation. In Proceedings of the IEEE conference on computer vision and pattern recognition (pp. 7472-7481).

\bibitem{Kim2020}Kim, M. and Byun, H., 2020. Learning texture invariant representation for domain adaptation of semantic segmentation. In Proceedings of the IEEE/CVF Conference on Computer Vision and Pattern Recognition (pp. 12975-12984).

\end{thebibliography}
\end{document}